\documentclass[10pt]{iopart}

\usepackage[utf8]{inputenc}
\usepackage{subcaption}
\usepackage[parfill]{parskip}
\usepackage{multirow}
\usepackage{blindtext}
\usepackage{color,soul}
\usepackage[normalem]{ulem}
\usepackage{graphicx}
\usepackage{xcolor}
\usepackage{url}

\usepackage{amsmath}
\usepackage{commath}
\usepackage[]{mathtools}
\usepackage{amsfonts}
\usepackage{stmaryrd}
\usepackage{algorithmic}
\usepackage{breqn}

\newcommand*{\argmin}{\operatorname{argmin}}

\begin{document}

\title{Federated Block-Term Tensor Regression for decentralised data analysis in healthcare}

\author{A. Faes$^{1, 2, *}$, Ashkan Pirmani$^{1, 2, 3}$, Yves Moreau$^2$, Liesbet M. Peeters$^{1, 2}$}
\address{$^1$ UHasselt, Biomedical Data Sciences, Data Science Institute, Hasselt, Belgium}
\address{$^2$ UHasselt, Biomedical Data Sciences, Immunology \& Infection, Biomedical Research Institute, Hasselt, Belgium}
\address{$^3$ KU Leuven, STADIUS, ESAT, Leuven, Belgium}
\eads{\mailto{axel.faes@uhasselt.be}}

\begin{abstract}
    Block-Term Tensor Regression (BTTR) has proven to be a powerful tool for modeling complex, high-dimensional data by leveraging multilinear relationships, making it particularly well-suited for applications in healthcare and neuroscience. However, traditional implementations of BTTR rely on centralized datasets, which pose significant privacy risks and hinder collaboration across institutions. To address these challenges, we introduce Federated Block-Term Tensor Regression (FBTTR), an extension of BTTR designed for federated learning scenarios. FBTTR enables decentralized data analysis, allowing institutions to collaboratively build predictive models while preserving data privacy and complying with regulations.

    FBTTR represents a major step forward in applying tensor regression to federated learning environments. Its performance is evaluated in two case studies: finger movement decoding from Electrocorticography (ECoG) signals and heart disease prediction. In the first case study, using the BCI Competition IV dataset, FBTTR outperforms non-multilinear models, demonstrating superior accuracy in decoding finger movements. For the dataset, for subject 3, the thumb obtained a performance of 0.76 $\pm$ .05 compared to 0.71 $\pm$ 0.05 for centralised BTTR. In the second case study, FBTTR is applied to predict heart disease using real-world clinical datasets, outperforming both standard federated learning approaches and centralized BTTR models. In the Fed-Heart-Disease Dataset, an AUC-ROC was obtained of 0.872 $\pm$ 0.02 and an accuracy of 0.772 $\pm$ 0.02 compared to 0.812 $\pm$ 0.003 and 0.753 $\pm$ 0.007 for the centralized model.

    Empirical evaluations highlight the robustness and scalability of FBTTR, achieving predictive performance comparable to centralized models while significantly reducing privacy concerns. Additionally, the algorithm is computationally efficient, making it viable for real-world deployment where timely predictions are critical. By offering a scalable and privacy-preserving solution, FBTTR has the potential to advance predictive analytics across multiple domains, including healthcare and brain-computer interface applications. Notably, FBTTR is developed as open-source software, promoting transparency and fostering collaboration within the research community.
\end{abstract}
% \submitto{\JNE}

\maketitle

\ioptwocol

\section{Introduction}
% - Motivation and significance of the study
% - Brief overview of tensor regression and its applications
% - Introduction to federated learning and its benefits in healthcare
% - Problem statement and objectives of the paper
\label{sec:introduction}
{Within regression analysis, multiway approaches are being applied more and more. As discussed in works such as} \cite{zhao2012higher, eliseyev2016penalized}, {they are used to model arm trajectories from Electrocorticography (ECoG) signals recorded from monkeys. Additionally, these approaches have been utilized in the context of exoskeleton-based arm trajectory, arm- and wrist rotations, and ECoG signals in tetraplegic patients, as highlighted in} \cite{benabid2019exoskeleton}. {Furthermore, multiway methods have been employed in the study of graded stimulation of the spinal cord region involved in walking in tetraplegic patients, as presented in} \cite{lorach2023walking}. {In the latter two scenarios, multiway partial least squares regression (NPLS) or a variation thereof is used within the decoders. However, this approach falls short in achieving the required level of accuracy for decoding fine finger movements, as discussed in} \cite{benabid2019exoskeleton}. {This limitation may be attributed to factors such as limited fitness capacity, high computational complexity, and slow convergence of NPLS, particularly when dealing with higher-order data, as highlighted in} \cite{zhao2012higher}. {On a more promising note, Zhao et al.} \cite{zhao2012higher} {have introduced a robust generalized framework known as Higher-Order Partial Least Squares (HOPLS). This framework is founded on a $(1, L_2, ..., L_N)$-rank Block Term Decomposition (BTD), where all blocks share the same multilinear rank, referred to as MTR (Multilinear Tensor Rank). HOPLS provides an optimal balance between fitness and model complexity, resulting in enhanced predictability. Consequently, HOPLS has demonstrated superior performance compared to conventional Partial Least Squares (PLS) approaches.}

{Camarrone and colleagues recently introduced Block-Term Tensor Regression (BTTR) as detailed in} \cite{camarrone2020accurate} and \cite{faes2022bttr}. {BTTR is grounded in Tucker decomposition and incorporates a deflation scheme that generates a sequence of blocks, each contributing successively less to the regression performance. Importantly, these blocks can possess varying MTR ranks, and their parameters are optimized automatically and on a block-by-block basis using Automatic Component Extraction (ACE). Notably, BTTR has demonstrated performance comparable to that of HOPLS, while significantly reducing training time. However, it is worth noting that, unlike HOPLS, BTTR is limited to predicting scalar variables. The aforementioned limitation becomes evident when decoding complex finger movements, which occur during actions such as grasping. A potential solution for avoiding this limitation is to employ an individual Block Term Tensor Regression (BTTR) model for each finger. However, this approach imposes constraints on the utilization of shared information among the fingers. Moreover, in situations where multiple fingers are engaged in simultaneous flexion, the recorded signals demonstrate temporal overlap and spatial sparsity, as discussed in} \cite{pan2018rapid}. {This presents a difficulty because exclusively training on individual finger movements might be inadequate for decoding the complexity of coordinated or multiple finger movements.}

Faes et al. recently proposed an extension of BTTR called eBTTR (extended BTTR) \cite{faes2022finger} to enable the prediction of coordinated finger flexions thereby also capturing finger co-activations, i.e.\ unintentional movements of other fingers. These co-activations are likely not encoded by the ECoG signal, but decoded by the eBTTR model as it is trained to replicate finger movements recorded with a data glove. However, as the recursive Tucker decomposition unfolds, some single-finger accuracies continue to improve but at the expense of other fingers, even to the extent that their accuracy becomes outperformed by that of (single-finger) BTTR. 

Federated learning (FL) is a relatively new machine learning paradigm, that addresses privacy and security concerns associated with centralized data storage and processing \cite{liu2024recent}. The concept of Federated Learning (FL) was introduced by McMahan et al. \cite{mcmahan2017communication}, where they proposed a method called Federated Averaging (FedAvg). This method aggregates locally computed updates from multiple devices to update a global model, reducing communication overhead and preserving data privacy. Bonawitz et al. \cite{bonawitz2019towards} extended this work by addressing secure aggregation of model updates, ensuring that individual updates remain private even in the presence of a partially trusted aggregator.

One of the primary motivations for FL is preserving data privacy. Geyer et al. \cite{geyer2017differentially} introduced a differentially private approach to FL, ensuring that the contribution of any single data point is obscured in the aggregated model updates. Abadi et al. \cite{abadi2016deep} explored deep learning with differential privacy, laying the groundwork for privacy-preserving techniques in FL. Truex et al. \cite{truex2019hybrid} proposed a hybrid approach combining secure multiparty computation (SMC) and differential privacy to enhance security in FL.

The scalability of FL systems is critical for practical deployment. Kairouz et al. \cite{kairouz2019advances} provided a comprehensive survey of advancements in FL, discussing techniques to improve the efficiency, scalability, and robustness of FL algorithms. Yang et al. \cite{yang2019federated} proposed a hierarchical FL framework to manage communication and computation load, making FL more scalable for large-scale deployments.

FL has been applied in various domains, demonstrating its versatility. Hard et al. \cite{hard2018federated} applied FL to mobile keyboard prediction, showing significant improvements in personalized model performance while preserving user privacy. Sheller et al. \cite{sheller2020federated} applied FL to medical imaging, demonstrating its potential in collaborative healthcare environments where data sharing is restricted due to privacy concerns.

Despite the progress, FL faces several challenges, including heterogeneity of data and devices, communication efficiency, and robustness against adversarial attacks. Li et al. \cite{li2020federated} provided a detailed analysis of these challenges and proposed potential solutions to address them. In summary, Federated Learning represents a significant shift towards decentralized machine learning, offering promising solutions for privacy-preserving data utilization. Ongoing research continues to address the technical challenges and expand the application scope of FL.

FL allows multiple parties to jointly train data-driven models while preserving data privacy - either by entirely avoiding data exchange or by means of data encryption \cite{nguyen2023towards}. Vertical Federated Learning (VFL) refers to scenarios where the private datasets share the same sample space but differ in the feature space. Cross-organizational processes naturally fit this scenario because even though local datasets can be mapped using the product or batch IDs, the local features are often distinct because each company usually operates a different type of sub-process. Although some research efforts have been devoted to adopting VFL in the field of process modeling, limited progress has been made.

In \cite{hartebrodt2021federated,grammenos2020federated}, the authors have proposed federated variants of Principal Component Analysis (PCA), which is a widely used method for fault detection and diagnosis. Nevertheless, in these studies, the application of PCA to process monitoring was not investigated. In \cite{duy2022towards}, Du et al. recently proposed a multiblock principal component analysis (MPCA) based federated multivariate statistical process control (FedMSPC) approach to jointly model a semiconductor fabrication process over multiple stages that are operated by different companies. Not only was the joint model more efficient (compared to the model trained on the downstream process only) for predicting process faults, but it also revealed important interactions between the two process stages that led to faulty batches in the downstream process. This study is among the first to demonstrate the prospects of (vertical) federated data analysis and process modeling for collaborative risk identification, decision-making, and value chain optimization.

However, as an unsupervised technique, FedMSPC's application is limited to scenarios where the relationship between process variables and some specific response variables (e.g., product quality, KPIs, etc.) is not of primary interest. Furthermore, previous methodology lacks mechanisms to incentivize data federation, which is pivotal to fostering collaboration among stakeholders along the value chain. In particular, participants should be rewarded based on their contribution to the overall outcomes (e.g., cost savings) resulting from data federation. Accessing contribution based merely on data quantity is certainly not enough, as one party may contribute a large volume of data that doesn't help much in solving the problem. Recent studies have proposed the use of the Shapley value in evaluating data contribution \cite{wang2019measure,wang2020principled}. However, since the Shapley value has an exponential time complexity, this approach poses significant inefficiency \cite{yang2023survey}

Millions of {individuals experience paralysis. This can be the result from a spinal cord injury due to, for instance, an accident. It can also occur from brainstem stroke, or from progressive disorders} like amyotrophic lateral sclerosis (ALS) \cite{SpinalCordInjury}. {Various assistive technologies have been proposed to provide an alternative means of communication} (for a review see \cite{ascari2018mobile}), {including brain computer interfaces (BCIs) used, among others, to select letters one-by-one to spell out words. Motor BCIs bypass the muscles of the human body and can thus be used to restore the function of, or replace those muscles. As discussed in} \cite{lorach2023walking}, {electrical stimulation can be an example of such a system. The stimulation occurs in the regions of the hand-knob area and the relevant spinal cord regions associated with walked. Additionally, BCIs contribute to the control of external limbs, such as prosthetic hands in case of full replacement, or exoskeletons in case of supporting existing function, or other effectors, as demonstrated in works such as} \cite{hochberg2006neuronal, collinger2013high, hochberg2012reach, wodlinger2014ten}. 

{Over the last few decades, electrocorticography (ECoG) has been gaining attention in the BCI research community, which involves electrodes placed on top of the cortical surface to record electrical activity. ECoG signals yield significantly higher amplitudes compared to scalp EEG, are not contaminated by artifacts, enjoy a broader bandwidth and higher spatial resolution\cite{miller2009power, staba2002quantitative}, as well as long-term signal stability \cite{wang2013electrocorticographic, nurse2017consistency}. When ECoG signals are recorded from the primary motor cortex, they reveal notable motor-related spatio-temporo-spectral patterns that can be leveraged to capture the dynamics of the corresponding movement, as discussed in works such as} \cite{anderson2012electrocorticographic, ball2009signal, bundy2016decoding}. {However, decoding rapid and coordinated finger movements, as in grasping and performing hand gestures, is challenging, calling for further algorithmic developments.}

Ramsey and {colleagues successfully classified four gestures based on hand motor Electrocorticography (ECoG) recordings in two subjects \cite{vansteensel2016fully}. These gestures were taken from the American Sign Language Alphabet. By extracting the local motor potentials (LMP) for each channel and using temporal template matching, classification was accomplished, as described in} \cite{bleichner2016give}. Li et al. \cite{li2017gesture} classified 3 gestures (“scissors,” “rock,” and “paper”) from hand motor ECoGs in 2 participants using an SVM-based classifier achieving 80\% accuracy on average in 3 participants. The classification results were also translated into commands for controlling a prosthetic hand in 2 participants. Pan et al. \cite{pan2018rapid} classified the same 3 gestures but used a recurrent neural network (RNN) that exploits the temporal information present in the ECoG recordings thereby achieving 90\% accuracy in 2 participants.

The prevalence of heart disease, one of the leading causes of mortality worldwide, necessitates accurate predictive models to aid in early diagnosis and treatment. Traditional machine learning models often require centralized data, which poses significant privacy risks in healthcare. Tensor regression models, such as the Block-Term Tensor Regression (BTTR), have shown promise in handling multi-dimensional data prevalent in medical records. However, the centralized nature of these models limits their scalability and applicability in real-world healthcare settings. Federated learning offers a solution to these challenges by enabling collaborative model training across multiple institutions without sharing sensitive data. Thus, in addition to the field of neuroscience, the proposed FBTTR framework can be applied to heart disease prediction, leveraging the advantages of tensor regression in a federated context to improve predictive accuracy while ensuring data privacy.

In this paper, we introduce Federated Block-Term Tensor Regression (FBTTR), an extension of BTTR designed for federated learning environments. Federated learning allows models to be trained across multiple decentralized institutions without sharing patient data, thus enhancing privacy. Our objective is to leverage the advantages of BTTR in a federated context to improve heart disease prediction while ensuring data privacy. Empirical evaluations demonstrate the robustness of FBTTR, showing that it achieves high predictive performance comparable to traditional centralized models. Moreover, the algorithm exhibits efficient computational performance, making it suitable for deployment in real-world healthcare environments where timely and accurate predictions are crucial. The results highlight FBTTR's potential in advancing predictive analytics in healthcare by providing a scalable, privacy-preserving solution for multiple domains including heart disease prediction and brain-computer interfaces.

\section{Methodology}
% - Overview of the block-term tensor regression algorithm
% - Detailed description of the Federated Block-Term Tensor Regression (FBTTR) framework
%   - Architecture and design
%   - Data partitioning and federation
%   - Model training and aggregation
%   - Privacy-preserving techniques
\label{sec:methodology}
\subsection{Block-Term Tensor Regression}
Block-Term Tensor Regression (BTTR) decomposes the input tensor into block components, facilitating efficient regression on high-dimensional data. It effectively captures the underlying structure of the data, making it suitable for applications involving multi-modal datasets \cite{faes2022bttr}. An overview of the mathematical notation used can be found in Table~\ref{math_not}. 

\begin{table}[!t]
\renewcommand{\arraystretch}{1.3}
\caption{Mathematical notation}
\label{math_not}
\centering
\begin{tabular}{|c p{14em}|}
\hline
Notation & Description \\
\hline
$\underline{\mathbf{T}}, \mathbf{M}, \mathbf{v}, S$ & tensor, matrix, vector, scalar (respectively)\\
$\mathbf{M}^T$ & transpose of matrix\\
$\times_n$ & mode-n product between tensor and matrix\\
$\otimes$ & Kronecker product\\
$\circ$ & outer product\\
$\norm{\cdot}_F$ & Frobenius norm\\
$\mathbf{T}_{(n)}$ & mode-n unfolding of tensor $\underline{\mathbf{T}}$\\
$\underline{\mathbf{C}}^{(T)}$ & core tensor associated to tensor $\underline{\mathbf{T}}$\\
$\mathbf{M}^{(n)}$ & mode-n factor matrix\\
$\mathbf{M}_{ind}$ & (sub-)matrix including the column(s) indicated in $ind$\\
$\mathbf{M}_{\backslash ind}$ & (sub-)matrix excluding the column(s) indicated in $ind$\\
$\llbracket \underline{\mathbf{C}} ; \mathbf{M}^{(1)},...,\mathbf{M}^{(N)} \rrbracket$ & full multilinear product $\underline{\mathbf{C}} \times_1 \mathbf{M}^{(1)} \times_2 \cdots \times_N \mathbf{M}^{(N)}$\\
$\langle \underline{\mathbf{T}} , \underline{\mathbf{E}} \rangle_{\{n,n\}}$ & mode-n cross-covariance tensor\\
\hline
\end{tabular}
\end{table}

The proposed regression model is based on the Block Term Decomposition (BTD) with automatic MTR determination, denoted as $(L^k_1, ..., L^k_N)$. This model employs a deflation-based approach to sequentially Tucker-decompose an ECoG tensor and a Finger Movement matrix into a sequence of blocks. Through the technique of Automatic Component Extraction (ACE), each block comprises representations maximally correlated.

Given a set of training data $\underline{\mathbf{X}}_{\text{train}} \in \mathbb{R}^{I_1 \times ... \times I_N}$ and a vectorial response $\mathbf{Y}_{\text{train}} \in \mathbb{R}^{I_1 \times M}$, training aims to automatically identify $\mathbf{K}$ blocks.

\begin{align*}
\begin{gathered}
    \underline{\mathbf{X}}_{\text{train}} = \sum^K_{k=1} \underline{\mathbf{G}}_{k} \times_1 \mathbf{t}_k \times_2 \mathbf{P}_k^{(2)} \times_3 ... \times_N \mathbf{P}_k^{(n)} + \underline{\mathbf{E}}_{k} \\
    \mathbf{Y}_{\text{train}} = \sum^K_{k=1} \mathbf{u}_k \mathbf{q}_k^T + \mathbf{F}_k \text{ with } \mathbf{u}_k = \mathbf{t}_k b_k
\end{gathered}
\end{align*}

Here, $\underline{\mathbf{G}}_{k} \in \mathbb{R}^{1 \times R_2^k \times ... \times R_N^k}$ represents the core tensor of the k-th block, $\mathbf{P}_k^{(n)}$ represents the loading matrix for the n-mode of the k-th block, $\mathbf{u}_k$ and $\mathbf{t}_k$ are the latent components, $\mathbf{q}_k$ is the loading matrix, $b_k$ is the regression coefficient, and $\underline{\mathbf{E}}_{k}$ and $\mathbf{F}_k$ represent residuals. Once the model is trained, and the values of $\underline{\mathbf{G}}_{k}$, $\mathbf{P}_k^{(n)}$, and $b_k$ are computed, the final prediction is obtained as follows: $\mathbf{Y}_{\text{test}} = \mathbf{TZ} = \mathbf{X}_{\text{test(1)}}\mathbf{WZ}$, where each column $\mathbf{w}_k = (\mathbf{P}_k^{(n)} \otimes ... \otimes \mathbf{P}_k^{(2)}) \text{vec}(\underline{\mathbf{G}}_{k})$ and each row $z_k = b_k \mathbf{q}_k$.

\subsection{Federated Learning Definition}

Federated learning (FL) is a learning paradigm in which multiple parties train collaboratively without the need to exchange or centralize data sets. A general formulation of FL reads as follows: Let \(L\) denote a global loss function obtained via a weighted combination of \(K\) local losses \( \{ L_k \}_{k=1}^K \), computed from private data \(X_k\), which is residing at the individual involved parties and never shared among them:
\begin{equation}
\min_{\phi} L(X, \phi) \quad \text{with} \quad L(X, \phi) = \sum_{k=1}^K w_k L_k(X_k, \phi),
\label{eq:1}
\end{equation}
where \(w_k > 0\) denote the respective weight coefficients.

In practice, each participant typically obtains and refines a global consensus model by conducting a few rounds of optimization locally before sharing updates, either directly or via a parameter server. The more rounds of local training are performed, the less it is guaranteed that the overall procedure is minimizing (Eq.~\ref{eq:1}). The actual process for aggregating parameters depends on the network topology, as nodes might be segregated into subnetworks due to geographical or legal constraints. There are many other ways to classify FL:

\begin{itemize}
    \item \textbf{Horizontal or vertical} - Depending on the partition of features in distributed datasets, FL can be classified as horizontal or vertical. Vertical Federated Learning is used for cases in which each device contains datasets with different features but from the same sample instances. For instance, two organizations that have data about the same group of people with different feature sets can use Vertical FL to build a shared ML model. Horizontal Federated Learning is used for cases in which each device contains datasets with the same feature space but with different sample instances. The first use case of FL - Google Keyboard - uses this type of learning in which the participating mobile phones have different training data with the same features.
    
    \item \textbf{Synchronous or asynchronous} - Depending on the aggregation strategy at an FL server, FL can be classified as synchronous or asynchronous. A synchronous FL server aggregates local models from a selected set of clients into a global model. An asynchronous FL server immediately updates the global model after a local model is received from a client, thereby reducing the waiting time and improving training efficiency.
    
    \item \textbf{Hub-and-spoke or peer-to-peer} - Aggregation strategies can rely on a single aggregating node (hub and spokes models) or on multiple nodes without any centralization. An example is peer-to-peer FL, where connections exist between all or a subset of the participants and model updates are shared only between directly connected sites, whereas an example of centralized FL aggregation is given in Algorithm~\ref{example}. Note that aggregation strategies do not necessarily require information about the full model update; clients might choose to share only a subset of the model parameters for the sake of reducing communication overhead, ensuring better privacy preservation or to produce multi-task learning algorithms having only part of their parameters learned in a federated manner.
\end{itemize}

\begin{figure}
\label{alg:fedavg}
\begin{algorithmic}[1]
    \renewcommand{\algorithmicrequire}{\textbf{Input:}}
    \renewcommand{\algorithmicensure}{\textbf{Output:}}
\REQUIRE num\_federated\_rounds \(T\)
% \PROCEDURE{Aggregating}{}
    \STATE Initialize global model: \(W(0)\)
    \FOR{\(t \leftarrow 1 \ldots T\)}
        \FOR{\(k \leftarrow 1 \ldots K \textbf{(in parallel)} \)} 
            \STATE Send \(W(t-1)\) to client \(k\)
            \STATE Receive model updates and number of local training iterations \((\Delta W_k^{(t-1)}, N_k)\) from client's local training with \(L_k(X_k, W^{(t-1)})\)
        \ENDFOR
        \STATE \(W^{(t)} \leftarrow W^{(t-1)} + \frac{1}{\sum_k N_k} \sum_k (N_k \Delta W_k^{(t-1)})\)
    \ENDFOR
    \STATE \RETURN \(W(t)\)
% \ENDPROCEDURE
\end{algorithmic} 
\caption{\label{example} Example of a FL algorithm via Hub \& Spoke (Centralized topology) with FedAvg aggregation} 
\end{figure}

\subsection{Federated Block-Term Tensor Regression (FBTTR)}
Federated Block-Term Tensor Regression (FBTTR) extends BTTR by integrating it into a federated learning framework. The architecture consists of multiple local models trained as seperate clients. These models are periodically synchronized by aggregating their parameters on a central server. The steps involved in FBTTR are:

\begin{itemize}
    \item \textbf{Data Partitioning and Federation} - The dataset is partitioned across multiple institutions, each retaining its data locally.
    \item \textbf{Local Model Training} - Each client trains a local BTTR model on its data.
    \item \textbf{Model Aggregation} - The local models' parameters are securely aggregated to update the global model.
\end{itemize}

The main aggregating node, the server, is responsible for coordinating the training process. The server initializes the global model and sends it to the clients. The clients train their local models and send the updates back to the server. The server aggregates the updates and updates the global model. This process is repeated for a specified number of federated rounds. The server then returns the final global model. The algorithm for FBTTR is shown in Figure~\ref{fbttr}. We also provide a detailed description of the local block in Figure~\ref{fbttr-block}. Typically, in standard BTTR, ACE uses the parameters SNR and $\tau$ to determine the number of components. In FBTTR, we extend this by introducing a "federated" version of ACE, where the parameters SNR and $\tau$ are optimized across multiple institutions. FBTTR needs to make sure that the dimensions of the tensor factorization is equivalent across all clients. Thus, an additional test is performed to ensure that the dimensions of the factor matrices are the same across all clients. If the dimensions are not the same, the server will send the correct dimensions to the clients.

In the end, the server contains the final model, which can be used for prediction. The server can also send the model back to the clients. This way, the clients can use the global model for prediction on their local data. The server can also send the model to new clients, who can use the model for prediction on their data. This way, the model can be used for prediction on new data. 

We can also see that the main distinct factor between FBTTR and other BTTR variations is the seperation of ACE and BTTR. Instead, first, ACE is used locally on each client and the server decides the number of components based on the local ACE results. This way, the server can ensure that the number of components is the same across all clients. This is important because the dimensions of the factor matrices need to be the same across all clients. If the dimensions are not the same, the server will send the correct dimensions to the clients. This way, the server can ensure that the dimensions of the factor matrices are the same across all clients. This also means that FBTTR incorporates horizontal federated learning, as the features are the same across all clients.

\begin{figure}
\begin{algorithmic}[1]
    \renewcommand{\algorithmicrequire}{\textbf{Input:}}
    \renewcommand{\algorithmicensure}{\textbf{Output:}}
\REQUIRE num\_federated\_blocks \(B\)
    \FOR {$k=1$ to $B$}
        % \COMMENT{Extract model parameters for this block using ACE:}
        \FOR{\(k \leftarrow 1 \ldots K \textbf{(in parallel)} \)}
            \STATE $\underline{\mathbf{G}}_{kb}^{(X)}, \mathbf{P}_{kb}^{(2)}, ..., \mathbf{P}_{kb}^{(N)}, \textit{SNR}, \tau $ = \textit{ACE}($\underline{\mathbf{E}}_0, \mathbf{F}_0$)
        \ENDFOR
        % \COMMENT{Local BTTR Block execution:}
        \STATE Find $\textit{SNR}_k$ and $\tau_k$ for client k such that all clients have the same dimensions
        \FOR{\(k \leftarrow 1 \ldots K \textbf{(in parallel)} \)} 
            \STATE Send $\textit{SNR}_k$ and $\tau_k$ to client \(k\)
            \STATE Receive model updates $\underline{\mathbf{G}}_{kb}^{(X)}, \mathbf{P}_{kb}^{(2)}, ..., \mathbf{P}_{kb}^{(N)}$ called \((\Delta W_k^{(t-1)}, N_k)\) from client's local training (see FBTTR Local Block).
        \ENDFOR
        % \COMMENT{Equivalent of FedAvg:}
        \STATE \(W^{(t)} \leftarrow \frac{1}{\sum_k N_k} \sum_k (N_k \Delta W_k^{(t-1)})\)
        % \COMMENT{Updating model parameters:}
        \FOR{\(k \leftarrow 1 \ldots K \textbf{(in parallel)} \)} 
            \STATE Send \(W(t-1)\) to client \(k\) and update the previous block \((\underline{\mathbf{E}}_k, \mathbf{F}_k)\)
        \ENDFOR
    \ENDFOR
    \STATE \RETURN \(W(t)\)
% \ENDPROCEDURE
\end{algorithmic} 
\caption{\label{fbttr} FBTTR server with FedAvg aggregation} 
\end{figure}

\begin{figure}
\begin{algorithmic}[1]
\renewcommand{\algorithmicrequire}{\textbf{Input:}}
\renewcommand{\algorithmicensure}{\textbf{Output:}}
\REQUIRE $\underline{\mathbf{X}} \in \mathbb{R}^{I_1 \times ... \times I_N}, \mathbf{Y} \in \mathbb{R}^{I_1 \times M}, k$
\ENSURE $\{ \mathbf{P}_k^{(n)} \}, \{ \mathbf{t}_k \}, \{ \mathbf{q}_k \}, \underline{\mathbf{G}}_k^{(X)}$ for $n=2,...,N$

\IF{$\norm{\underline{\mathbf{E_k}}} > \epsilon$ and $\norm{\mathbf{F_k}} > \epsilon$}
    \STATE $\underline{\mathbf{C_{k}}} = \langle \underline{\mathbf{E}}_k , \mathbf{F}_k \rangle_{\{1,1\}}$

    % \STATE $\underline{\mathbf{G}}_k^{(X)}, \mathbf{q}_k, \mathbf{t}_k, \mathbf{P}_k^{(2)}, ..., \mathbf{P}_k^{(N)} $ = \textit{ACE}($\underline{\mathbf{E}}_k, \mathbf{F}_k$)
    \STATE $\underline{\mathbf{G}}_k, \mathbf{q}_k, \{ \mathbf{P}_k^{(n)} \}^N_{n=2}$ = \textit{F-mPSTD}($\underline{\mathbf{E}}_k$, $\mathbf{F}_k$, $\textit{SNR}^*$, $\tau^*$) 
    \STATE $\mathbf{t}_k = (\underline{\mathbf{E}}_k \times_2 \mathbf{P}_k^{(2)T} \times_3 ... \times_N \mathbf{P}_k^{(N)T})_{(1)}\textit{vec}(\underline{\mathbf{G}}_k^{(X)})$
    \STATE $\mathbf{t}_k = \mathbf{t}_k / \norm{\mathbf{t}_k}_F$
    \STATE $\underline{\mathbf{G}}_k^{(X)} = \llbracket \underline{\mathbf{E}}_k ; \mathbf{t}_k^T, \mathbf{P}_k^{(2)T},...,\mathbf{P}_k^{(N)T} \rrbracket$

    \STATE $\mathbf{u_k} = \mathbf{F_k} \mathbf{q_k}$
    \STATE $\mathbf{d_k} = \mathbf{u_k}^T \mathbf{t_k}$

    \COMMENT{Deflation:}

    \STATE $\underline{\mathbf{E_{k+1}}} = \underline{\mathbf{E_{k}}} - \llbracket \underline{\mathbf{G}}_k^{(X)} ; \mathbf{t}_k, \mathbf{P}_k^{(2)},...,\mathbf{P}_k^{(N)} \rrbracket$
    \STATE $\mathbf{F_{k+1}} = \mathbf{F_{k}} - \mathbf{d_k} \mathbf{t_k} \mathbf{q_k}^T$
\ELSE
    \STATE \textbf{continue}
\ENDIF

\end{algorithmic} 
\caption{\label{fbttr-block} FBTTR Local Block} 
\end{figure}

\subsubsection{Automatic Component Extraction (ACE)}

Given an N-way variable $\underline{\mathbf{X}} \in \mathbb{R}^{I_1 \times ... \times I_N}$ and a vectorial variable $\mathbf{Y} \in \mathbb{R}^{I_1 \times M}$, we aim to automatically extract the latent components $\mathbf{t}$, $\mathbf{q}$ and ${P^{(n)} }_{(n=2)}^N$, associated with the n-th mode of $\underline{\mathbf{X}}$ and maximally correlated with $\mathbf{Y}$, while $\norm{\underline{\mathbf{X}} - \llbracket \underline{\mathbf{G}} ; \mathbf{t}, \mathbf{P}^{(2)},...,\mathbf{P}^{(N)} \rrbracket}_F$ is minimized. 

\begin{figure}
\begin{algorithmic}[1]
\renewcommand{\algorithmicrequire}{\textbf{Input:}}
\renewcommand{\algorithmicensure}{\textbf{Output:}}
\REQUIRE $\underline{\mathbf{X}} \in \mathbb{R}^{I_1 \times ... \times I_N}, \mathbf{Y} \in \mathbb{R}^{I_1 \times M}$
\ENSURE $\underline{\mathbf{G}}^{(X)} \in \mathbb{R}^{1 \times R_2 \times ... \times R_N}, \mathbf{q}, \mathbf{t}, \{ \mathbf{P}^{(n)} \}^N_{n=2}$

\STATE $\underline{\mathbf{C}} = \langle \underline{\mathbf{X}} , \mathbf{Y} \rangle_{(1)}$
\STATE \textbf{Initialisation of} $\tau = 90,...,100$; \textit{SNR}$=1,...,50$

\FOR {$\textit{SNR}_i$ in \textit{SNR}}
    \FOR {$\tau_j$ in $\tau$}
        \STATE $\underline{\mathbf{G}}, \mathbf{q}, \{ \mathbf{P}^{(n)} \}^N_{n=2}$ = \textit{F-mPSTD}($\underline{\mathbf{X}}$, $\mathbf{Y}$, $\textit{SNR}_i$, $\tau_j$) 
        \STATE \textbf{calculate} BIC value corresponding to $\textit{SNR}_i$ and $\tau_j$ using Eq~\ref{BIC}
    \ENDFOR
    \STATE \textbf{select} $\tau^*$ = $\argmin_{\tau} \text{BIC}(\tau)$
    \STATE \textbf{calculate} BIC value corresponding to $\textit{SNR}_i$ and $\tau^*$ using Eq~\ref{BIC}
\ENDFOR

\STATE \textbf{select} $\textit{SNR}^*$ = $\argmin_{\textit{SNR}} \text{BIC}(\textit{SNR}, \tau^*)$
\STATE $\underline{\mathbf{G}}, \mathbf{q}, \{ \mathbf{P}^{(n)} \}^N_{n=2}$ = \textit{F-mPSTD}($\underline{\mathbf{X}}$, $\mathbf{Y}$, $\textit{SNR}^*$, $\tau^*$) 
\STATE $\mathbf{t} = (\underline{\mathbf{X}} \times_2 \mathbf{P}^{(2)T} \times_3 ... \times_N \mathbf{P}^{(N)T})_{(1)}\textit{vec}(\underline{\mathbf{G}})$
\STATE $\mathbf{t} = \mathbf{t} / \norm{\mathbf{t}}_F$
\STATE $\underline{\mathbf{G}}^{(X)} = \llbracket \underline{\mathbf{X}} ; \mathbf{t}^T, \mathbf{P}^{(2)T},...,\mathbf{P}^{(N)T} \rrbracket$

\RETURN $\underline{\mathbf{G}}^{(X)}, \mathbf{q}, \mathbf{t}, \{ \mathbf{P}^{(n)} \}^N_{n=2}$, $\textit{SNR}^*$, $\tau^*$
\end{algorithmic} 
\caption{\label{ace} ACE} 
\end{figure}

Within ACE, we define the mode-1 cross-product between predictor and response variables as $\underline{\mathbf{C}} = \langle \underline{\mathbf{X}} , \mathbf{Y} \rangle_{(1)}$ and its decomposition as $\underline{\mathbf{C}} \approx \llbracket \underline{\mathbf{G}}^{(c)} ; \mathbf{q}, \mathbf{P}^{(2)},...,\mathbf{P}^{(N)} \rrbracket$. We provide the model with automatic SNR and $\tau$ selection based on Bayesian Information Criterion (BIC) defined here as:

\begin{multline}
BIC(\tau, \text{SNR} | \text{SNR}, \tau^*) = \\ 
\log( \frac{\norm{\underline{\mathbf{C}} - \llbracket \underline{\mathbf{G}}^{(c)} ; \mathbf{q}, \mathbf{P}^{(2)},...,\mathbf{P}^{(N)} \rrbracket}_F}{s}) + \frac{\log(s)}{s} DF, 
\label{BIC}
\end{multline}

where $\underline{\mathbf{G}}^{(c)}$, $q$ and $\{\mathbf{P}^{(n)}\}^N_{n=2}$ are the sparse core, latent vector and factor matrices obtained with F-mPSTD \cite{faes2022bttr} - a modified version of the original  sparse Tucker decomposition (PSTD) \cite{yokota2014multilinear} - using specific $\tau$ and SNR values, $s$ the number of entries in $\underline{\mathbf{G}}$, and $DF$ the degree of freedom calculated as the number of non-zero elements in $\underline{\mathbf{G}}^{(c)}$, as suggested in \cite{allen2011sparse}. For each SNR value, the associated optimal $\tau$ is computed as $\tau^* = \argmin_\tau BIC(\tau, \text{SNR}) $. Then, the optimal SNR is determined as $\text{SNR}^* = \argmin_{\text{SNR}} BIC(\text{SNR}, \tau^*) $. Once $\underline{\mathbf{G}}^{(c)}$, $q$ and $\{\mathbf{P}^{(n)}\}^N_{n=2}$ are computed, the score vector $t$ is first calculated as

\begin{align*}
    \mathbf{t} = (\underline{\mathbf{C}} \times_2 \mathbf{P}^{(2)T} \times_3 ... \times_N \mathbf{P}^{(n)T})_{(1)} \text{vec}(\underline{\mathbf{G}}^{(c)}),
\end{align*}

and then normalized. This is summarized in Algorithm~\ref{ace}.
    
\begin{figure}
\begin{algorithmic}[1]
\renewcommand{\algorithmicrequire}{\textbf{Input:}}
\renewcommand{\algorithmicensure}{\textbf{Output:}}
\REQUIRE $\underline{\mathbf{X}} \in \mathbb{R}^{I_1 \times ... \times I_N}, \mathbf{Y} \in \mathbb{R}^{I_1 \times M}, \tau, $\textit{SNR}
\ENSURE $\underline{\mathbf{G}} \in \mathbb{R}^{1 \times R_2 \times ... \times R_N}, \mathbf{q}, \{ \mathbf{P}^{(n)} \}^N_{n=2}$
\\ \textit{Initialisation} :
\STATE $\underline{\mathbf{C}} = \langle \underline{\mathbf{X}} , \mathbf{y} \rangle_{(1)} \in \mathbb{R}^{1 \times I_2 \times ... \times I_N}$
\STATE \textbf{Initialisation of} $\{ \mathbf{P}^{(n)} \}^N_{n=2}$, $\mathbf{q}$ \text{and} $\underline{\mathbf{G}}$ \text{using HOOI on} $\underline{\mathbf{C}}$
\\ \textit{LOOP Process}
\REPEAT
    \STATE \textbf{update} $\underline{\mathbf{G}}$ using \textit{SNR}
    \STATE \textbf{prune} $\{ \mathbf{P}^{(n)} \}^N_{n=2}$, $\mathbf{q}$ and $\underline{\mathbf{G}}$ using $\tau$
\UNTIL{convergence is reached}
\RETURN $\underline{\mathbf{G}}, \mathbf{q}, \{ \mathbf{P}^{(n)} \}^N_{n=2}$ 
\end{algorithmic} 
\caption{\label{mpstd} F-mPSTD} 
\end{figure}

The F-mPSTD model is first initialized with higher-order orthogonal iteration (HOOI) \cite{de2000best}. Then, iteratively, a soft-thresholding rule based on parameter $\lambda$, alternated with a threshold $\tau$, are applied to enhance model sparsity and to prune irrelevant components, respectively. Note that in \cite{yokota2014multilinear} SNR $\in [1, 50]$ is used to derive, via a line search, the optimal degree of sparsity $\lambda$ of the core tensor (see \cite{yokota2014multilinear}. At each iteration, the core tensor $\underline{\mathbf{G}}$ is updated using the soft-thresholding rule as $\underline{\mathbf{G}} = sgn(\underline{\mathbf{G}}) \times max\{\abs{\underline{\mathbf{G}}} - \lambda, 0\}$, while the threshold $\tau \in [0, 100]$ is used to reject unnecessary components from the n-mode $S^{(n)} = \{r { | } 100 (1 - \frac{\sum_i{\mathbf{G}_{(n) (r, i)}}}{ \sum_{t,i}{\mathbf{G}_{(n) (t, i)}} }) \ge \tau\}$, $\mathbf{P}^{(n)} = \mathbf{P}^{(n)} (:, S^{(n)})$, $\mathbf{q} = \mathbf{q} (S^{(n)})$ and $\mathbf{G}^{(n)} = \mathbf{G}^{(n)} (S^{(n)}, :)$. The F-mPSTD is summarized in Algorithm~\ref{mpstd}. 

\subsection{Non-multilinear approaches}

To provide a comprehensive comparison of the performance of the Federated Block Term Tensor Regression (FBTTR) method with other established and more recent techniques, we will employ the BCI competition IV dataset (details provided later). Specifically, we will benchmark FBTTR against the competition-winning method, which utilized a linear regression model based on amplitude modulation (AM) \cite{liang2012decoding}. Additionally, we will compare it with more contemporary approaches that have leveraged Random Forests (RF), Convolutional Neural Networks (CNN), and Long Short-Term Memory Networks (LSTM), as proposed and evaluated in \cite{xie2018decoding}. Notably, all of these models fall under the category of non-multilinear approaches.

However, it is important to note that in both \cite{liang2012decoding} and \cite{xie2018decoding}, only a single test set was used to evaluate model performance. This methodological limitation restricts the possibility of performing statistically rigorous significance testing between the competing algorithms. To overcome this limitation and ensure robust comparisons, we have re-implemented the AM, RF, LARS, CNN, and LSTM models. Our approach ensures a fair and reliable performance evaluation across multiple test sets. The detailed steps for each implementation are described below.

For the AM-based method, we followed the procedure outlined in \cite{liang2012decoding} meticulously. First, the ECoG signals were filtered into three distinct frequency bands: sub-gamma (1-60Hz), gamma (60-100Hz), and high-gamma (100-200Hz). For each of these bands, we computed the amplitude modulation (AM) and derived their respective band-specific AM features. These features were then subject to a feature selection process, where forward feature selection using a wrapper approach was applied to identify the most relevant AM features for each individual subject and finger. The selected features were subsequently used to fit a linear regression model. Finally, we assessed the performance of the model and compared it against the results originally reported in \cite{liang2012decoding} to ensure consistency.

For the LARS model, we employed the {\it LassoLars} function from the most recent version of the {\it scikit-learn} Python package (version 0.24.2, released in April 2021). Although Xie et al. did not explicitly state the version they used, we assume they used version 0.19.1, as that would have been current during their research. In our implementation, LARS first transforms the original signal using Independent Component Analysis (ICA), decomposing the signal into different frequency bands. From these bands, the band powers were computed, and a LassoLars regression model was subsequently fitted. The primary tuning parameter for LARS is $\alpha$, which determines the weight of the penalty term, where $\alpha = 0$ corresponds to ordinary least squares (OLS) regression. We employed a line search technique to optimize the $\alpha$ parameter for best performance.

Similarly, for the Random Forest (RF) model, we adopted an analogous preprocessing pipeline. We used the {\it RandomForestRegressor} function from the same version of the {\it scikit-learn} package. Like LARS, RF also performs signal decomposition using ICA and band power calculation. The processed features were then used to train the {\it RandomForestRegressor} model.

For the CNN and LSTM models, we followed the architectural specifications and training procedures provided in \cite{xie2018decoding}. The CNN model applies a linear regression model to the features extracted using a convolutional neural network, whereas the LSTM model uses a recurrent network to capture temporal dependencies in the data. Both models were implemented using the PyTorch framework in Python (version 1.8.0, released in March 2021), although Xie et al. did not specify the version of PyTorch used in their experiments.

By re-implementing these methods and testing them on multiple datasets, we aim to provide a more robust comparison, overcoming the limitations of previous single-test-set studies.

\section{Results and Discussion}

\subsection{Case Study: Brain-Computer Interfacing Finger Movement Decoding}
We will compare FBTTR's performance with that of the aforementioned non-multilinear models, as well as with multilinear models HOPLS and BTTR, in predicting continuous finger flexions from ECoG recordings. We will use the publicly-available BCI Competition IV dataset 4 for reproducibility's sake. It comprises ECoG signals sampled at 1000 Hz from the motor cortex (hand-knob area) of three subjects, as well as the time courses of the flexion of each of five fingers of the contralateral hand (measured with a data glove). There are 150 trials (samples) in total (30 trials per finger) recorded in a single session (600 seconds). Subjects flexed the cued finger 3-5 times for 2 seconds followed by a rest period of 2 seconds. The first 400 and the last 200 seconds of recording were used as training and testing sets, respectively. The number of ECoG electrodes (channels) was 62 for Subject 1, 48 for Subject 2, and 64 for Subject 3. More details about the data can be found in \cite{miller2008prediction}.

The dataglove position measurement lags by 37ms ($\pm$ 3ms, SEM) the amplifier measurement. However, this is of the same order of granularity as the dataglove's position measurement as it is sampled at 25Hz, thus every 40ms. Hence, the dataglove position measurement is shifted by 1 position in order to account for the lag. This is also done in the approaches against which we compare our (F)BTTR's performance.

The glove data and the ECoG signals were extracted starting 1 second prior to trial onset (``epoch''). ECoG recordings were prepared first by filtering out the power line using notch filters centered at 50 and 100 Hz, then, by removing bad channels (i.e., those exhibiting unstable or unchanging signals, i.e., channel 55 in Subject 1, channels 21 and 38 in Subject 2, and channel 50 in Subject 3). The remaining channels were re-referenced to a Common Average Reference (CAR) [3]: the average of all signals is taken as reference and subtracted from all signals \cite{binnie2003clinical}. The ECoG signals are transformed into a 4th-order ECoG tensor $\underline{\mathbf{X}} \in  \mathbb{R}^{\text{Samples} \times \text{Channels} \times \text{Frequencies} \times \text{Time}}$ as follows:

\begin{itemize}
    \item \textbf{Samples} depends on the length of the data glove's trajectory vector $Y$.
    \item \textbf{Channels} corresponds to the number of curated electrodes (thus, after removing bad ones), i.e., 61 for Subject 1, 46 for Subject 2, and 63 for Subject 3.
    \item \textbf{Frequencies} corresponds to the 8 bidirectional fourth-order Butterworth band-pass filters \cite{tucker1966some} ECoG signals are subjected to extract the corresponding spectral amplitudes in the $\delta$ (1.5-5 Hz), $\theta$ (5-8 Hz), $\alpha$ (8-12 Hz), $\beta_1$ (12-24 Hz), $\beta_2$ (24-34 Hz), $\gamma_1$ (34-60 Hz), $\gamma_2$ (60-100 Hz), $\gamma_3$ (100-130 Hz) bands.
    \item \textbf{Time} is composed of 10 instances or bins as for each of the described 8 components, the most recent 1 second epoch is downsampled to 10 Hz.
\end{itemize}

The dataglove data is normalized (z-scored) independently for each finger, yielding a vector $Y \in  \mathbb{R}^{\text{Samples} \times \text{Fingers}}$. \textbf{Samples} corresponds to the sampled finger flexions over time. The \textbf{Fingers} dimension corresponds to the 5 fingers: Thumb, Index, Middle, Ring, and Pinky.

As a result, for each channel $c$, a 3rd-order ECoG tensor $\underline{\mathbf{X}}_c \in  \mathbb{R}^{\text{Samples} \times \text{Frequencies} \times \text{Time}}$ is computed after epoch selection, band-pass filtering, and 10 Hz downsampling. These steps are repeated for each time sample $s$. The results are then merged into a matrix $\mathbf{X}_{c,f} \in \mathbb{R}^{\text{Samples} \times \text{Time (10 bins)}}$. This matrix is then normalized (z-scored) to reduce the difference in magnitude between frequency bands: $x_{s,t} = (x_{s,t} - \mu_{c,f} )/ \delta_{c,f}$ where $\mu_{c,f}$ and $\delta_{c,f}$ are, respectively, the mean and standard deviation computed for channel $c$ and frequency band $f$ of the training set. Note that the same values are used to normalize the test set.

Once all $\mathbf{X}_{c,f} \in \mathbb{R}^{\text{Samples} \times \text{Time (10)}}$ are computed for the various components (8), they are merged into $\underline{\mathbf{X}}_c \in  \mathbb{R}^{\text{Samples} \times \text{Frequencies (8)} \times \text{Time (10)}}$.

Next, when all 3rd-order ECoG tensors $\underline{\mathbf{X}}_c \in  \mathbb{R}^{\text{Samples} \times \text{Frequencies (8 components)} \times \text{Time (10 bins)}}$ are computed for each channel $c$, they are merged into $\underline{\mathbf{X}} \in  \mathbb{R}^{\text{Samples} \times \text{Channels \{61, 46, 63\}} \times \text{Frequencies (8)} \times \text{Time (10)}}$.

\subsubsection{Parameter optimization multilinear models, performance assessment}

In order to optimize the model parameters from the training data, ${K, R_2,...,R_N}$ for HOPLS and K for BTTR, eBTTR and FBTTR, a 5-fold cross-validation approach was used. 

Since the cited BCI Competition IV dataset 4 studies reported Pearson's correlation coefficients between data glove- and predicted intended finger trajectories, we also applied it here. In support of the statistical analysis, the test data was split in 5 non-overlapping blocks. We used the two-tailed Wilcoxon signed-rank test \cite{wilcoxon1992individual} to compare average accuracies per finger and subject; they are considered significantly different if the p-value is $<0.05$. 

\subsubsection{Results and Impact}

The Pearson correlation coefficients are reported for the intended movements of each finger individually, and averaged across all fingers except for finger 4 (ring) as flexing the latter is difficult to suppress when the 3rd or 5th finger is flexing, as this was done by the authors of the cited papers. The average results and their standard deviations for the 5 blocks of test data (see above) are listed in Table~\ref{results-subj-1}, Table~\ref{results-subj-2} and Table~\ref{results-subj-3} (listed under Avg.) for subjects 1, 2 and 3, respectively. 

\begin{figure*}[!htb]
    \includegraphics[width=\textwidth]{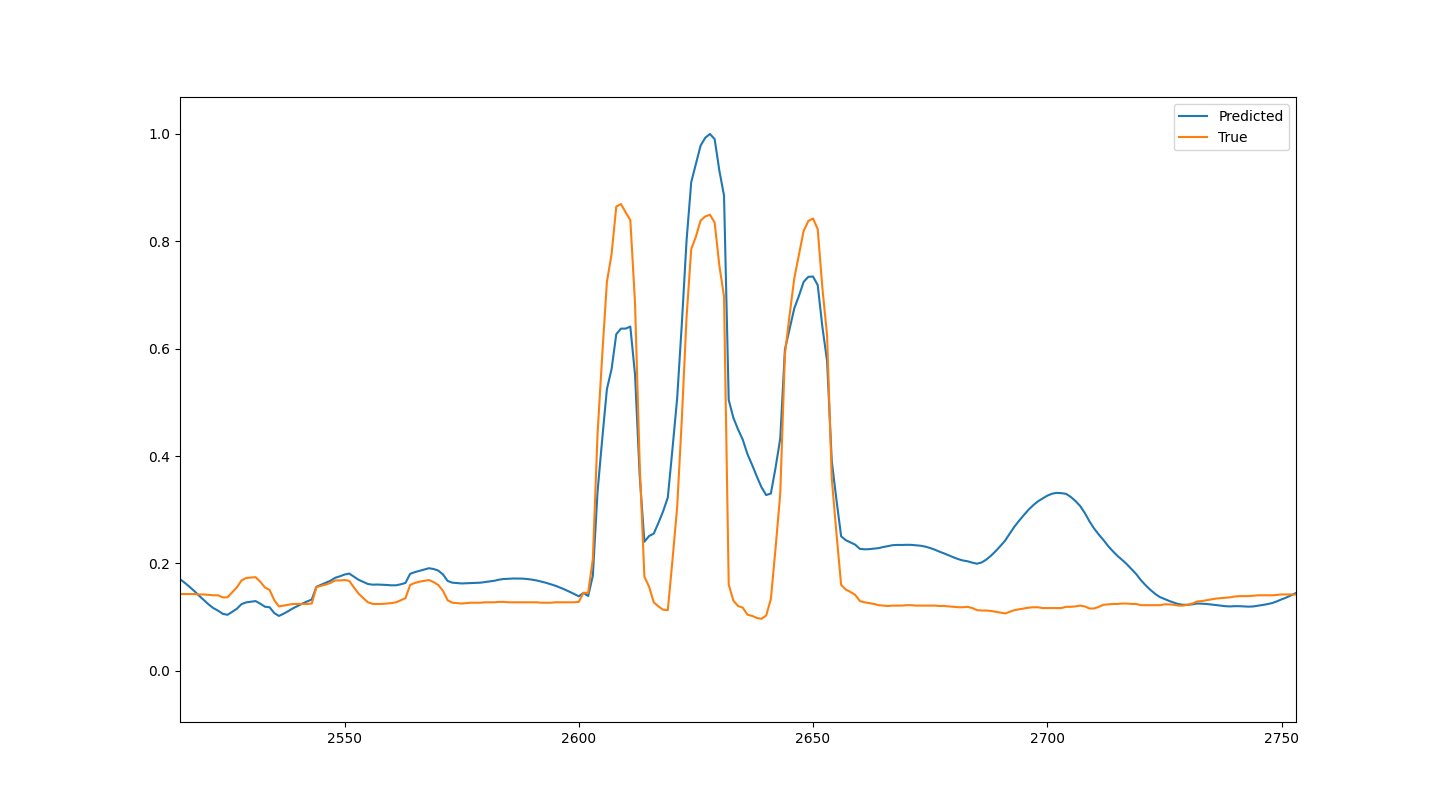}
    \caption{\label{scheme}Predicted ring finger flexions of Subject 1. Actual (data glove) and predicted (regression models) finger flexion amplitudes (z-scores) are plotted as function of time (in ms), Curve colors are explained in the inset. For the sake of exposition, only results for the FBTTR are shown.} 
\end{figure*}

A statistically significant difference between FBTTR and HOPLS for subjects 1 and 2, mainly due to the difference in correlation coefficients of the middle finger. Interestingly, there is no significant performance difference between FBTTR and eBTTR for subject 1 and subject 2, but there is for subject 3. This is due to the difference in correlation coefficients of the thumb. Thus, overall, we can say there is a small, but significant improvement of using FBTTR over eBTTR. Figure~\ref{scheme} shows an example of predicted finger movement for FBTTR together with the data glove signal.

\begin{figure*}[!htb]
    \includegraphics[width=\textwidth]{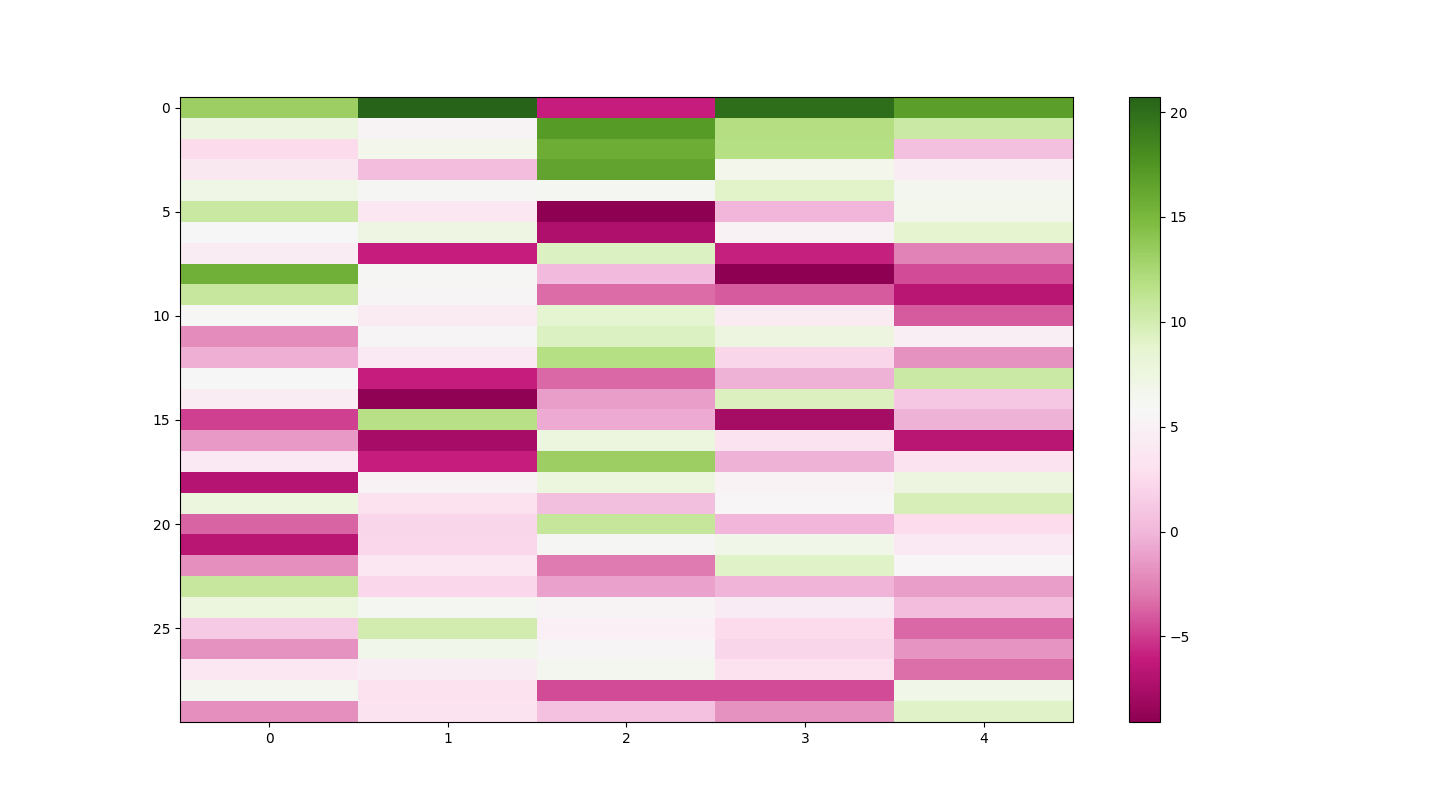}
    \caption{\label{model}Model parameters of Subject 1 for the first 30 blocks. The x-axis represents the fingers (Thumb, Index, Middle, Ring and Pinky respectively), while the y-axis represents the various blocks in order. The color intensity represents the strength of the connection in terms of Pearson correlation.}
\end{figure*}

Looking at the model parameters for subject 1 (Figure~\ref{model}), we can see that there is generally a smooth transition from block to block for each finger and thus, the transition between these blocks is quite stable over time. This tells us that the model is learning the underlying structure of the data well. It also tells us that the FedAVG aggregation method is working well, as the model parameters are quite stable over time and not causing any significant fluctuations. 

We also investigated the runtime needed to estimate the model parameters. One of the main advantages of BTTR is its speed. BTTR is trained significantly faster than HOPLS (e.g. ~3 minutes against ~14 hours for HOPLS). Note that HOPLS requires computationally expensive techniques such as cross-validation to identify the optimal set of model parameters (i.e., the number of scores and loadings). Since FBTTR is based on BTTR, it enjoys the latter's fast training time (e.g. ~5 minutes against ~3 minutes for BTTR). Averaging the model parameters doesn't impact the runtime significantly. However, there are likely some performance improvements that can be made. Generally, it's not required to train the model from scratch for each new client. Instead, the model can be fine-tuned using a few epochs of training on the client's data. This is a common practice in federated learning, where the global model is fine-tuned on the client's data to adapt to the local data distribution. This fine-tuning process is much faster than training the model from scratch.

\begin{table}
\begin{center}
\resizebox{\columnwidth}{!}{%
\begin{tabular}{ |c|c|c|c|c|c|c|c| } 
    \hline
    Methods & Thumb & Index & Middle & Ring & Pinky & Avg. \\
    \hline
    F-BTTR & 0.72 $\pm$ .06 & 0.77 $\pm$ .09 & 0.45 $\pm$ .02 & 0.68 $\pm$ .04 & 0.67 $\pm$ .02 & 0.64 $\pm$ .05 \\
    eBTTR & 0.71 $\pm$ .04 & 0.75 $\pm$ .06 & 0.49 $\pm$ .06 & 0.69 $\pm$ .04 & 0.68 $\pm$ .01 & 0.66 $\pm$ .03 \\
    HOPLS & 0.70 $\pm$ .05 & 0.79 $\pm$ .08 & \textbf{0.36 $\pm$ .03} & 0.70 $\pm$ .06 & 0.65 $\pm$ .02 & 0.63 $\pm$ .04 \\
    AM & \textbf{0.57 $\pm$ .03} & 0.69 $\pm$ .06 & \textbf{0.14 $\pm$ .02} & \textbf{0.52 $\pm$ .04} & \textbf{0.28 $\pm$ .01} & \textbf{0.42 $\pm$ .03} \\
    RF & \textbf{0.58 $\pm$ .09} & \textbf{0.54 $\pm$ .05} & \textbf{0.07 $\pm$ .03} & \textbf{0.31 $\pm$ .05} & \textbf{0.33 $\pm$ .02} & \textbf{0.38 $\pm$ .05} \\
    LARS & \textbf{0.11 $\pm$ .05} & \textbf{0.08 $\pm$ .03} & \textbf{0.10 $\pm$ .02} & 0.60 $\pm$ .05 & \textbf{0.39 $\pm$ .02} & \textbf{0.17 $\pm$ .03}  \\
    CNN & \textbf{0.67 $\pm$ .04} & 0.78 $\pm$ .04 & \textbf{0.11 $\pm$ .02} & \textbf{0.54 $\pm$ .03} & \textbf{0.45 $\pm$ .04} & \textbf{0.50 $\pm$ .04} \\
    LSTM & 0.73 $\pm$ .03 & 0.79 $\pm$ .08 & \textbf{0.18 $\pm$ .02} & 0.61 $\pm$ .04 & \textbf{0.45 $\pm$ .04} & \textbf{0.54 $\pm$ .04} \\
    \hline
\end{tabular}
}
\caption{\label{results-subj-1} Intended (cued) finger movement accuracy for Subject 1 (Pearson correlation).}
\end{center}
\end{table}

\begin{table}
\begin{center}
\resizebox{\columnwidth}{!}{%
\begin{tabular}{ |c|c|c|c|c|c|c|c| } 
    \hline 
    Methods & Thumb & Index & Middle & Ring & Pinky & Avg. \\
    \hline
    F-BTTR & 0.66 $\pm$ .06 & 0.47 $\pm$ .09 & 0.32 $\pm$ .02 & 0.52 $\pm$ .04 & 0.48 $\pm$ .02 & 0.48 $\pm$ .05 \\
    eBTTR & 0.63 $\pm$ .05 & 0.47 $\pm$ .08 & 0.33 $\pm$ .03 & 0.52 $\pm$ .02 & 0.47 $\pm$ .01 & 0.48 $\pm$ .05 \\
    HOPLS & 0.63 $\pm$ .04 & 0.47 $\pm$ .06 & \textbf{0.26 $\pm$ .05} & 0.51 $\pm$ .02 & 0.48 $\pm$ .01 & 0.44 $\pm$ .04 \\
    AM & \textbf{0.52 $\pm$ .03} & \textbf{0.36 $\pm$ .06} & \textbf{0.23 $\pm$ .02} & 0.48 $\pm$ .04 & \textbf{0.33 $\pm$ .01} & \textbf{0.36 $\pm$ .03} \\
    RF & \textbf{0.52 $\pm$ .05} & \textbf{0.36 $\pm$ .04} & \textbf{0.22 $\pm$ .03} & \textbf{0.39 $\pm$ .04} & \textbf{0.25 $\pm$ .02} & \textbf{0.34 $\pm$ .04}  \\
    LARS & \textbf{0.54 $\pm$ .05} & 0.41 $\pm$ .04 & \textbf{0.18 $\pm$ .02} & \textbf{0.44 $\pm$ .04} & \textbf{0.25 $\pm$ .02} & \textbf{0.35 $\pm$ .03} \\
    CNN & 0.60 $\pm$ .04 & 0.40 $\pm$ .04 & \textbf{0.24 $\pm$ .02} & \textbf{0.44 $\pm$ .03} & \textbf{0.28 $\pm$ .04} & \textbf{0.38 $\pm$ .04} \\
    LSTM & 0.62 $\pm$ .03 & 0.38 $\pm$ .08 & 0.27 $\pm$ .02 & 0.47 $\pm$ .04 & \textbf{0.30 $\pm$ .04} & \textbf{0.39 $\pm$ .04} \\
    \hline
\end{tabular}
}
\caption{\label{results-subj-2} Idem to Table \ref{results-subj-1} but for Subject 2.}
\end{center}
\end{table}
\begin{table}
\begin{center}
\resizebox{\columnwidth}{!}{%
\begin{tabular}{ |c|c|c|c|c|c|c|c| } 
    \hline
    Methods & Thumb & Index & Middle & Ring & Pinky & Avg. \\
    \hline 
    FBTTR & 0.76 $\pm$ .05 & 0.57 $\pm$ .06 & 0.64 $\pm$ .02 & 0.62 $\pm$ .02 & 0.76 $\pm$ .01 & 0.68 $\pm$ .05 \\
    eBTTR & \textbf{0.71 $\pm$ .05} & 0.57 $\pm$ .07 & 0.64 $\pm$ .04 & 0.62 $\pm$ .02 & 0.73 $\pm$ .01 & 0.66 $\pm$ .05 \\
    HOPLS & 0.74 $\pm$ .06 & 0.57 $\pm$ .09 & 0.65 $\pm$ .02 & 0.61 $\pm$ .04 &  \textbf{0.68 $\pm$ .02} & 0.64 $\pm$ .04 \\
    AM & \textbf{0.59 $\pm$ .03} & 0.51 $\pm$ .06 & \textbf{0.32 $\pm$ .02} & \textbf{0.53 $\pm$ .04} & \textbf{0.42 $\pm$ .01} & \textbf{0.46 $\pm$ .03} \\
    RF & 0.67 $\pm$ .05 & \textbf{0.27 $\pm$ .04} & \textbf{0.16 $\pm$ .03} & \textbf{0.14 $\pm$ .04} & \textbf{0.36 $\pm$ .02} & \textbf{0.37 $\pm$ .04} \\
    LARS & 0.72 $\pm$ .05 & \textbf{0.43 $\pm$ .04} & \textbf{0.45 $\pm$ .02} & \textbf{0.51 $\pm$ .04} & \textbf{0.64 $\pm$ .02} & \textbf{0.56 $\pm$ .03} \\
    CNN & 0.74 $\pm$ .03 & 0.53 $\pm$ .05 & \textbf{0.45 $\pm$ .04} & \textbf{0.49 $\pm$ .03} & 0.68 $\pm$ .06 & \textbf{0.60 $\pm$ .05} \\
    LSTM & 0.74 $\pm$ .02 & 0.55 $\pm$ .06 & \textbf{0.46 $\pm$ .04} & \textbf{0.41 $\pm$ .02} & 0.75 $\pm$ .06 & \textbf{0.62 $\pm$ .05} \\
    \hline
\end{tabular}
}
\caption{\label{results-subj-3} Idem to Table \ref{results-subj-1} but for Subject 3.}
\end{center}
\end{table}

\subsection{Case Study: Heart Disease Prediction}
Our dataset for this case study comprises two real-world clinical datasets from Flamby \cite{ogier2022flamby}: Fed-Heart-Disease and Fed-Tcga-Brca. The Fed-Heart-Disease dataset contains 740 records from four centers, detailing 13 clinical features and a binary heart disease indicator for each patient. The Fed-Tcga-Brca dataset, sourced from The Cancer Genome Atlas's Genomics Data Commons portal, has data from 1,088 breast cancer patients across six centers. With 39 clinical features and each patient's time of death.

Our experimental setup involves training the FBTTR model on federated data partitions. We evaluate the model using metrics such as accuracy, precision, recall, and F1-score, comparing its performance against centralized BTTR and other baseline models. To illustrate the effectiveness of FBTTR, we follow different experimental scenarions as described in the FL4E framework \cite{pirmani2024accessible}. Foundational foundational scripts that can be generalized and act as guidelines for hybrid experiment settings in the FL4E framework were used for these experiments.

\begin{itemize}
    \item \textbf{Fully Federated Experiment}: All participating clients contribute to the global model using their local datasets, adhering to the conventional FL setup.
    \item \textbf{Hybrid Experiment}: Some clients contribute their data centrally while others participate as federated clients. This scenario demonstrates how partial centralization can influence model performance.
    \item \textbf{Centralized Experiment}: All clients send their data to a central server for model training. This serves as a benchmark to compare the federated and hybrid approaches against a fully centralized model.
\end{itemize}

Although FBTTR, for now, only implements the equivalent for the FedAVG algorithm, we still made the comparison with other federated learning strategies, such as FedOpt and FedProx, to evaluate its performance in a federated setting \cite{reddi2020adaptive, li2020federated2}. 

\subsubsection{Parameter optimization multilinear models, performance assessment}

In order to optimize the model parameters from the training data, $K$ for BTTR, eBTTR and FBTTR, a 5-fold cross-validation approach was used. Hyperparameter tuning was performed for standard federated learning using grid search on a centralized setting and applied to each experiment of the datasets. To ensure consistency, each experiment was repeated five times. 

While Pearson Correlation's are the standard for finger movement decoding, this dataset is evaluated using Area Under the Receiver Operating Characteristic Curve (ROC-AUC) and accuracy metrics for the Fed-Heart-Disease dataset and the concordance index (C-index) for the Fed-Tcga-Brca dataset. These metrics were chosen according to their ability to assess model performance in binary classification and survival analysis task, while accounting for potential class imbalance. Similar to the other case study, in support of the statistical analysis, the test data was split in 5 non-overlapping blocks. We used the two-tailed Wilcoxon signed-rank test \cite{wilcoxon1992individual} to compare average accuracies; they are considered significantly different if the p-value is $<0.05$. 

\subsubsection{Results and Impact}

The case study results highlight the practical applicability of FBTTR in predicting heart disease. The implementation demonstrated that FBTTR could be a viable solution for predictive analytics in healthcare. Similarly as in the other case study, we found that federated models exhibit similar to superior performance compared to centralized models. This is conform to results found in the literature \cite{pirmani2024accessible}. The hybrid experiment also showed promising results, indicating that a combination of centralized and federated learning can be beneficial in certain scenarios and can be a potential avenue for FBTTR. 

The results for the Fed-Heart-Disease Dataset are shown in Table~\ref{fed-heart-disease}. A first observation is that FBTTR outperforms all other federated learning strategies, including FedAvg, FedAdagrad, FedYogi, and FedProx, as well as the hybrid and centralized experiments, in absolute numbers, with most comparisons being statistically significant. 

While there is a slight accuracy drop in the federated setting compared to the centralized setting, the difference is not statistically significant, and there is practically no difference in the ROC-AUC. 

The results for the Fed-Tcga-Brca Dataset are shown in Table~\ref{fed-tcga-brca}. Here, FBTTR also outperforms all other federated learning strategies, including FedAvg, FedAdagrad, FedYogi, and FedProx, as well as the hybrid and centralized experiments, in absolute numbers, with most comparisons being statistically significant. The only comparisons that are not significant are between FBTTR and FedAdagrad, and FBTTR and FedProx. A possible explanation for this is that the FedAdagrad and FedProx algorithms are more robust to the noise in the data, which could explain why they perform better compared to the other federated learning strategies in this case. 

FBTTR also performs statistically significantly better than BTTR in the Fed-TCGA-BRCA Datasets. This is an important result, as it shows that FBTTR is able to leverage the federated data to improve the prediction performance compared to a centralized model. 

\begin{table}[h]
\centering
\caption{\label{fed-heart-disease} Experiments Using FL4E (\cite{pirmani2024accessible}) and FBTTR with the Fed-Heart-Disease Dataset. Significantly different results compared FBTTR are indicated in bold.}
\begin{tabular}{|l|c|c|}
\hline
\textbf{Method}  & \textbf{ROC-AUC}           & \textbf{Accuracy}       \\ \hline
\textbf{FBTTR}          & 0.872 $\pm$ 0.02       & 0.772 $\pm$ 0.02           \\ \hline
\textbf{BTTR}            & 0.874 $\pm$ 0.03       & 0.783 $\pm$ 0.04          \\ \hline
\textbf{Fully Federated} &                        &                       \\ \hline
FedAvg           & 0.846 $\pm$ 0.002               & 0.733 $\pm$ 0.007          \\ \hline
FedAdagrad       & 0.841 $\pm$ 0.029               & \textbf{0.726 $\pm$ 0.05}           \\ \hline
FedYogi          & 0.803 $\pm$ 0.051               & \textbf{0.715 $\pm$ 0.032}          \\ \hline
FedProx          & 0.846 $\pm$ 0.003               & 0.741 $\pm$ 0.006          \\ \hline
\textbf{Hybrid Experiment} &                        &                       \\ \hline
FedAvg           & 0.825 $\pm$ 0.004               & 0.740 $\pm$ 0.054          \\ \hline
FedAdagrad       & 0.821 $\pm$ 0.008               & 0.741 $\pm$ 0.005          \\ \hline
FedYogi          & 0.794 $\pm$ 0.013               & \textbf{0.710 $\pm$ 0.039}          \\ \hline
FedProx          & 0.822 $\pm$ 0.004               & \textbf{0.737 $\pm$ 0.012}          \\ \hline
\textbf{Local Experiment}  &                        &                       \\ \hline
Client 0         & 0.842 $\pm$ 0.009               & 0.753 $\pm$ 0.011          \\ \hline
Client 1         & 0.882 $\pm$ 0.007               & 0.800 $\pm$ 0.013          \\ \hline
Client 2         & 0.546 $\pm$ 0.271               & \textbf{0.550 $\pm$ 0.199}          \\ \hline
Client 3         & 0.542 $\pm$ 0.054               & \textbf{0.559 $\pm$ 0.096}          \\ \hline
\textbf{Centralized}       & 0.812 $\pm$ 0.003       & 0.753 $\pm$ 0.007          \\ \hline
\end{tabular}
\end{table}

\begin{table}[h]
\centering
\caption{\label{fed-tcga-brca} Experiments Using FL4E (\cite{pirmani2024accessible}) and FBTTR with the Fed-TCGA-BRCA Dataset. Significantly different results compared FBTTR are indicated in bold.}
\begin{tabular}{|l|c|}
\hline
\textbf{Method}  & \textbf{C-Index}       \\ \hline
\textbf{FBTTR}          & 0.775 $\pm$ 0.01          \\ \hline
\textbf{BTTR}            & \textbf{0.737 $\pm$ 0.02}          \\ \hline
\textbf{Fully Federated} &                \\ \hline
FedAvg           & \textbf{0.732 $\pm$ 0.030}          \\ \hline
FedAdagrad       & 0.748 $\pm$ 0.016          \\ \hline
FedYogi          & 0.745 $\pm$ 0.037          \\ \hline
FedProx          & \textbf{0.725 $\pm$ 0.007}          \\ \hline
\textbf{Hybrid Experiment} &                \\ \hline
FedAvg           & \textbf{0.656 $\pm$ 0.06}           \\ \hline
FedAdagrad       & 0.776 $\pm$ 0.036          \\ \hline
FedYogi          & \textbf{0.726 $\pm$ 0.041}          \\ \hline
FedProx          & \textbf{0.439 $\pm$ 0.227}          \\ \hline
\textbf{Local Experiment}  &                \\ \hline
Client 0         & \textbf{0.668 $\pm$ 0.064}          \\ \hline
Client 1         & \textbf{0.445 $\pm$ 0.237}          \\ \hline
Client 2         & \textbf{0.635 $\pm$ 0.166}          \\ \hline
Client 3         & \textbf{0.570 $\pm$ 0.140}          \\ \hline
Client 4         & \textbf{0.851 $\pm$ 0.078}                \\ \hline
Client 5         & \textbf{0.666 $\pm$ 0.001}          \\ \hline
\textbf{Centralized}       & \textbf{0.609 $\pm$ 0.207}          \\ \hline
\end{tabular}
\end{table}
        
\section{Conclusion}
In the first case study, using the BCI Competition IV dataset, we compared FBTTR with non-multilinear models for finger movement decoding. Our results show that FBTTR outperforms the other models, demonstrating its effectiveness in predicting finger movements from ECoG signals. It performs on par, and in some cases, better compared to centralised versions of BTTR. In the second case study, we applied FBTTR to predict heart disease using two real-world clinical datasets. Our findings indicate that FBTTR performs well in this domain, showcasing its potential for healthcare analytics other than finger movement decoding. The second usecase also shows the superior performance of FBTTR compared to standard federated learning, as well as the equal or superior performance compared to BTTR. Overall, our results suggest that FBTTR is a versatile and powerful tool for a wide range of applications, from brain-computer interfacing to predictive healthcare analytics.

FBTTR represents a significant advancement in applying tensor regression in federated learning environments. The method's ability to handle high-dimensional data while preserving privacy makes it particularly suited for healthcare applications. However, there are limitations, such as the computational overhead associated with model synchronization and the potential for performance degradation with highly heterogeneous data. Future research could explore optimizing the aggregation process and extending the method to other medical conditions.

In conclusion, this paper introduced Federated Block-Term Tensor Regression (FBTTR), an innovative approach that extends BTTR to federated learning settings. Our findings demonstrate that FBTTR effectively predicts heart disease and reconstruct finger movements, making it a valuable tool for healthcare analytics. Future work will focus on refining the model and exploring its application in other domains.

\section*{Acknowledgements}
We acknowledge the support from [Funding Sources], and thank our collaborators at [Institutions] for their valuable contributions to this research.

\bibliographystyle{unsrt}
\bibliography{main}

\end{document}